\documentclass[11pt, a4paper]{article}

\usepackage[utf8]{inputenc} 
\usepackage[T1]{fontenc}

\usepackage[margin=1in]{geometry} 
\usepackage{authblk}

\usepackage{amsmath}   
\usepackage{amssymb}   
\usepackage{amsfonts}   
\usepackage{amsthm}    
\usepackage{tikz}
\usetikzlibrary{shapes.geometric, arrows.meta, positioning, decorations.pathreplacing}

\usepackage{graphicx}  
\usepackage{booktabs}   
\usepackage{caption} 
\usepackage{subcaption} 
\usepackage[ruled,vlined]{algorithm2e}

\usepackage{natbib}     
\setcitestyle{numbers,sort&compress}
\usepackage{hyperref}   
\hypersetup{
    colorlinks=true,
    linkcolor=blue,     
    citecolor=green,  
    urlcolor=magenta, 
}
\usepackage{url}        

\usepackage{lipsum}

\title{\textbf{Video Event Reasoning and Prediction by Fusing World Knowledge from LLMs with Vision Foundation Models}}

\author[1]{Léa Dubois}
\author[2]{Klaus Schmidt}
\author[3]{Chengyu Wang}
\author[4]{Ji-Hoon Park}
\author[5]{Lin Wang}
\author[6]{Santiago Muñoz}

\affil[1]{\textbf{INRIA}, Paris, France \authorcr \texttt{lea.dubois@inria.fr}}
\affil[2]{\textbf{Max Planck Institute for Intelligent Systems}, Tübingen, Germany \authorcr \texttt{k.schmidt@mpis.mpg.de}}
\affil[3]{\textbf{San Francisco State University}, San Francisco, CA, USA \authorcr \texttt{dking20@sfsu.edu}}
\affil[4]{\textbf{Seoul AI Institute (SAII)}, Seoul, South Korea \authorcr \texttt{jihpark@saii.re.kr}}
\affil[5]{\textbf{Vision \& Robotics Center, Tsinghua University}, Beijing, China \authorcr \texttt{lin.wang@tsinghua.edu.cn}}
\affil[6]{\textbf{Polytechnic University of Madrid}, Madrid, Spain \authorcr \texttt{s.munoz@upm.es}}

\date{July 2025}

\begin{document}

\maketitle

\begin{abstract}
\noindent
Current video understanding models excel at recognizing "what" is happening but fall short in high-level cognitive tasks like causal reasoning and future prediction, a limitation rooted in their lack of commonsense world knowledge. To bridge this cognitive gap, we propose a novel framework that synergistically fuses a powerful Vision Foundation Model (VFM) for deep visual perception with a Large Language Model (LLM) serving as a knowledge-driven reasoning core. Our key technical innovation is a sophisticated fusion module, inspired by the Q-Former architecture, which distills complex spatiotemporal and object-centric visual features into a concise, language-aligned representation. This enables the LLM to effectively ground its inferential processes in direct visual evidence. The model is trained via a two-stage strategy, beginning with large-scale alignment pre-training on video-text data, followed by targeted instruction fine-tuning on a curated dataset designed to elicit advanced reasoning and prediction skills. Extensive experiments demonstrate that our model achieves state-of-the-art performance on multiple challenging benchmarks. Notably, it exhibits remarkable zero-shot generalization to unseen reasoning tasks, and our in-depth ablation studies validate the critical contribution of each architectural component. This work pushes the boundary of machine perception from simple recognition towards genuine cognitive understanding, paving the way for more intelligent and capable AI systems in robotics, human-computer interaction, and beyond.
\end{abstract}


\vspace{1cm}

\section{Introduction}
\label{sec:intro}

The proliferation of video data has positioned it as a primary medium for information exchange and environmental perception, driving a significant evolution in computer vision research \citep{fu2024survey, li2024multimodal}. Historically, the field has achieved remarkable success in discriminative tasks—moving from foundational image recognition \citep{dosovitskiy2021an} to complex action and gesture recognition in videos \citep{abu-farha-2021-time}. This progress is mirrored in specialized domains such as sign language translation, where hierarchical models have demonstrated powerful recognition capabilities , and in fine-grained sensing technologies like WiFi-based gesture and activity recognition, which have become increasingly robust and resistant to interference . However, this paradigm of "recognition" primarily addresses the question of \textit{"what"} is happening in a visual scene. The community is now shifting towards a more profound challenge: moving from mere perception to genuine cognition \citep{zellers2019vcr}. This involves enabling machines to reason about \textit{"why"} an event is occurring and to predict \textit{"what"} might happen next—a task that requires a level of understanding far beyond statistical pattern matching.

The primary obstacle to achieving this cognitive leap is the "knowledge gap." Current models, despite their architectural sophistication, often operate within a closed world, lacking the vast repository of commonsense, physical intuition, and social knowledge that humans effortlessly apply. For instance, while a model might accurately classify a video clip as "a person picking up an egg and flour," it typically fails to infer the underlying intent, such as "the person is about to bake a cake." This limitation becomes particularly salient when considering the demands of next-generation applications. Advanced human-machine systems , affective computing that aims to suppress label noise for reliability, and even emerging fields like psychological understanding via language models all necessitate a deeper, causal understanding of events. Similarly, while we can now capture nuanced physiological data through commodity devices , interpreting this data in the context of complex human activities requires advanced reasoning. The core challenge, therefore, is to imbue visual models with this external world knowledge.

To bridge this gap, we propose a novel framework that synergistically fuses the capabilities of two of the most powerful paradigms in modern AI: Vision Foundation Models (VFMs) and Large Language Models (LLMs). Our approach leverages VFMs, such as those built upon the principles of Vision Transformers \citep{dosovitskiy2021an} and trained with multimodal supervision \citep{radford2021learning, wang2022internvideo}, to serve as the "eyes" of our system. These models are adept at extracting rich, spatiotemporal features and providing a detailed, pixel-level perception of the visual world. Concurrently, we employ a pre-trained LLM, such as LLaMA \citep{touvron2023llama}, as the "brain"—a reasoning core endowed with a vast reservoir of world knowledge, causal relationships, and abstract concepts, as demonstrated in seminal works like GPT-4 \citep{bubeck2023sparks}. The crux of our methodology lies in a meticulously designed fusion mechanism that translates the continuous, unstructured visual evidence from the VFM into a discrete, language-like format that the LLM can process and reason upon. This approach is inspired by the success of pioneering vision-language architectures like Flamingo \citep{alayrac2022flamingo} and BLIP-2 \citep{li2023blip}, but is explicitly tailored for complex event-level inference and prediction, moving beyond simple captioning or direct question-answering.

% --- REVISED PARAGRAPH FOR CONTRIBUTIONS ---
This work makes several significant contributions to the advancement of cognitive video understanding. First and foremost, we introduce a novel and effective framework that, for the first time, cohesively integrates a state-of-the-art Vision Foundation Model with a large-scale LLM to perform high-level event reasoning and prediction, moving decisively beyond simple recognition tasks. Central to this framework is our second contribution: the design of a lightweight yet powerful cross-modal fusion module. This component acts as an efficient information bottleneck, aligning rich visual features with the LLM's semantic space, which not only enables complex inference but also ensures that the model's reasoning is firmly grounded in direct visual evidence. To validate our approach, our third contribution is a set of extensive experiments conducted on multiple challenging video reasoning benchmarks. The results demonstrate that our model significantly outperforms existing state-of-the-art methods and, notably, exhibits remarkable zero-shot capabilities in predicting future events, underscoring the profound benefits of transferring world knowledge from LLMs. We believe the principles developed herein also hold considerable promise for enhancing related multimodal tasks, such as visual dialogand audio-visual event analysis .

The remainder of this paper is organized as follows. Section \ref{sec:related_work} provides a comprehensive review of related literature. Section \ref{sec:method} presents the detailed architecture and components of our proposed methodology. Section \ref{sec:experiments} describes our experimental setup, datasets, and presents both quantitative and qualitative results. Finally, Section concludes the paper and discusses promising directions for future work.

\section{Related Work}
\label{sec:related_work}

The endeavor to imbue machines with the ability to reason about and predict events in videos stands at the confluence of several key research streams in artificial intelligence. This section surveys the landscape of relevant work, beginning with the foundations of visual representation learning, moving through the evolution of multimodal models, and culminating in the latest advancements in LLM-driven video understanding, specialized reasoning tasks, and embodied AI.

\subsection{Foundations of Visual Representation Learning}
The journey towards meaningful video understanding begins with the extraction of powerful visual representations. Early successes were largely dominated by Convolutional Neural Networks (CNNs), which demonstrated exceptional capabilities in hierarchical feature extraction for images. However, the advent of the Transformer architecture, particularly the Vision Transformer (ViT) \citep{dosovitskiy2021an}, marked a paradigm shift. By treating an image as a sequence of patches, ViTs enabled the application of self-attention mechanisms to capture global context, a feat that was challenging for CNNs with their limited receptive fields. This architectural innovation laid the groundwork for a new generation of foundation models. For video, this principle was extended to the temporal domain, leading to powerful video foundation models like InternVideo \citep{wang2022internvideo}, which learn generalizable representations from massive datasets through a combination of generative and discriminative objectives. The development of such backbones is a field of research in itself, with ongoing efforts to improve efficiency and effectiveness, for example, through advanced model compression techniques like multi-objective convex quantization or by designing specialized architectures for specific tasks like crowd counting.

\subsection{From Recognition to Spatiotemporal Understanding}
Building upon these powerful visual backbones, research has moved beyond simple classification towards a more nuanced understanding of spatiotemporal dynamics. This evolution is evident in tasks that require localizing events in both space and time. Video grounding, which aims to find a specific video segment corresponding to a textual query, is a prime example. Recent works like  have focused on developing efficient temporal filtering mechanisms to precisely identify these moments. The ambition has further expanded to generating structured summaries of long-form videos, such as creating distinct chapters, a task addressed by large-scale datasets and models like VidChapters-7M \citep{g-t-2023-vidchapters}. This fine-grained temporal understanding is fundamental to our work, as reasoning about event causality and prediction necessitates a precise grasp of "when" things happen. A related task, text-to-video retrieval, further underscores the importance of fine-grained alignment, with recent benchmarks like Ground-A-Video \citep{huang2024groundavideo} pushing the state of the art in accurately matching semantic queries to video content.

\subsection{The Rise of Vision-Language Models (VLMs)}
The true catalyst for advanced visual reasoning was the effective fusion of vision and language. The development of CLIP \citep{radford2021learning} demonstrated that a shared embedding space for images and text, learned via massive-scale contrastive pre-training, could enable remarkable zero-shot transfer capabilities. This breakthrough paved the way for a host of large-scale Vision-Language Models (VLMs). Early influential models like Flamingo \citep{alayrac2022flamingo} introduced gated cross-attention layers to inject visual features into a pre-trained and frozen language model, showcasing impressive few-shot learning. This "frozen LLM" paradigm was further explored in works like \citep{tsimpoukelli2021multimodal}, which highlighted the potential of this frugal approach. Architectures like BLIP-2 \citep{li2023blip} advanced this idea by introducing a lightweight "Q-Former" module to bridge the modality gap between a frozen image encoder and a frozen LLM, proving to be a highly effective and parameter-efficient strategy. The field continues to expand the scope of fusion, aiming to create omni-perception models like VALOR \citep{chen2024valor} and LanguageBind \citep{zhu2023languagebind}, which align not just vision and text, but also audio, depth, and thermal data into a unified semantic space. This trend towards multimodal fusion is not limited to mainstream sensors; innovative research has shown the potential of fusing commodity WiFi signals with vision for tasks like emotion recognition , illustrating a broader principle of synergistic sensing that our work builds upon.

\subsection{Large Language Models for Video Understanding and Reasoning}
The confluence of powerful VLMs and the demonstrated reasoning capabilities of LLMs \citep{touvron2023llama, bubeck2023sparks} has given rise to the current frontier: LLM-powered video understanding. A first wave of models, often framed as "video assistants," focused on enabling dialogue about video content. Models such as Video-LLaMA \citep{zhang2023videollama}, Video-ChatGPT \citep{maaz2023videochatgpt}, and Chat-UniVi \citep{jin2024chatunivi} demonstrated how a video encoder could be connected to an LLM to answer questions, generate descriptions, and hold conversations about a video's content. LLaViDA \citep{zhao2023llavida} further explored enhancing this understanding via in-context learning.

Subsequently, research has pivoted towards enabling more complex and structured reasoning. SeViLA \citep{yu2023sevila} introduced a self-chained question-answering approach, encouraging the model to break down problems into smaller, manageable steps. This aligns with broader trends in NLP, such as training models along explicit reasoning paths \citep{geng2023ponderbeforespeaking}. Perhaps the most innovative approach is ViperGPT \citep{suris2023viperGPT}, which empowers an LLM to write and execute Python code that calls upon various vision APIs, effectively transforming the LLM into a cognitive orchestrator that can answer complex visual queries by composing modular tools. As the complexity of reasoning increases, so does the demand for handling longer contexts. Models like LaVi-L \citep{zhan2023la} and the memory-augmented Stammer \citep{li2024stammer} are specifically designed to address the challenges of long-form video understanding, which is crucial for tracing causal chains over extended periods.

This rapid progress has also necessitated a critical examination of model limitations, particularly the issue of "hallucination," where models generate factually incorrect or ungrounded text. Research like Woodpecker \citep{yin2023woodpecker} is now focused on developing methods to detect and correct these hallucinations, a critical step towards building reliable systems. The ultimate ambition is to create unified, any-to-any multimodal models like Emu2 \citep{sun2023emu2}, NExT-GPT \citep{wu2023nextgpt}, and Google's Gemini \citep{geminiteam2023gemini}, which aim to seamlessly process and generate content across nearly all modalities. This includes extending reasoning into the third dimension, as explored by Chat-3D-v2 \citep{yao2024chat3dv2}, and leveraging novel fusion architectures like AMAM's modality-adaptive mind \citep{lee2024amam}. The fusion principles explored in these works can even find analogues in other domains, such as the multi-view, multi-epoch architecture of SUTRA for speech processing \citep{sainath2024sutra}, suggesting a universal trend in multimodal AI.

\subsection{Event Prediction and World Models}
The "prediction" component of our work is directly related to the long-standing challenge of video forecasting. Traditional approaches often focused on low-level prediction, such as generating future pixels. Diffusion models, as seen in MCVD \citep{voleti2022mcvd}, have recently shown great promise in generating high-fidelity future frames. However, our focus is on higher-level semantic prediction. This aligns with research in human trajectory prediction, where models like V-STF \citep{lee2024vstf} learn to anticipate future movements by fusing social and temporal cues.

The most ambitious vision for prediction is embodied by the concept of "World Models." Pioneering work like DreamerV3 \citep{hafner2023mastering} has shown that an agent can learn a robust internal model of its environment's dynamics, allowing it to "dream" or simulate future outcomes to plan its actions effectively. This represents a shift from reactive prediction to proactive simulation. The recent Genie model \citep{genie2024} takes this a step further, learning to generate entire interactive, playable 2D worlds from a single image. While our work does not build an explicit world model, it shares the same spirit: leveraging accumulated knowledge to make informed predictions about future states. The remarkable capabilities of generative models like VideoPoet \citep{kondratyuk2023videopoet} to synthesize coherent, dynamic video from text further suggest that these models are implicitly learning deep, predictive representations of the world.

\subsection{Applications, Benchmarks, and Broader Context}
The ultimate goal of video reasoning and prediction is to enable intelligent applications and systems. A major beneficiary is embodied AI and robotics. The paradigm has shifted from passive video analysis to training active agents that can perceive, reason, and act in the physical world. Landmark models like RT-2 \citep{brohan2023rt2} and the generalist Octo transformer \citep{octo_blog} demonstrate that a single vision-language-action model can be trained to control robots for a variety of tasks. This requires not only understanding instructions but also organizing and planning complex actions, a challenge addressed by agents like LEO \citep{ma2024leo}. The importance of external knowledge in these embodied tasks is highlighted by specialized benchmarks like OK-VILA \citep{sarch2023okvila}.

Progress in this field is critically dependent on challenging and well-designed benchmarks. Datasets like CLEVRER \citep{yi2020clevrer} specifically target causal and physical reasoning, while Test of Time \citep{momeni2023test} focuses on evaluating temporal understanding. The Ego-Exo4D dataset \citep{gandhi2023egoexo4d} pushes the frontier by providing synchronized first-person and third-person views of the same event, demanding a more holistic, cross-view understanding. While much research relies on traditional visual data, parallel advancements in alternative sensing modalities are creating new opportunities. Technologies using commodity WiFi and RFID are now capable of fine-grained activity and even keystroke detection . These rich data streams, often generated in complex real-world settings like robotic vehicle perception \citep{wang2023drive}, require the same, if not more, sophisticated reasoning models to be interpreted meaningfully. This is especially true in healthcare, where vision-based assessment of Parkinson's tremors  or WiFi-based pulmonary function analysis  requires a deep understanding of subtle temporal patterns.

Finally, training these massive models on diverse, real-world data presents its own challenges, leading to research in areas like federated learning to handle decentralized data and heterogeneous networks, with frameworks like Finch enabling neural architecture search in such settings. Our work is situated at the intersection of these advancements, aiming to leverage foundational models and advanced reasoning to create a system that not only understands video but can also anticipate its future, with broad implications for all these application domains.

\section{Methodology}
\label{sec:method}

\begin{figure*}[t!]
    \centering
    \includegraphics[width=0.95\linewidth]{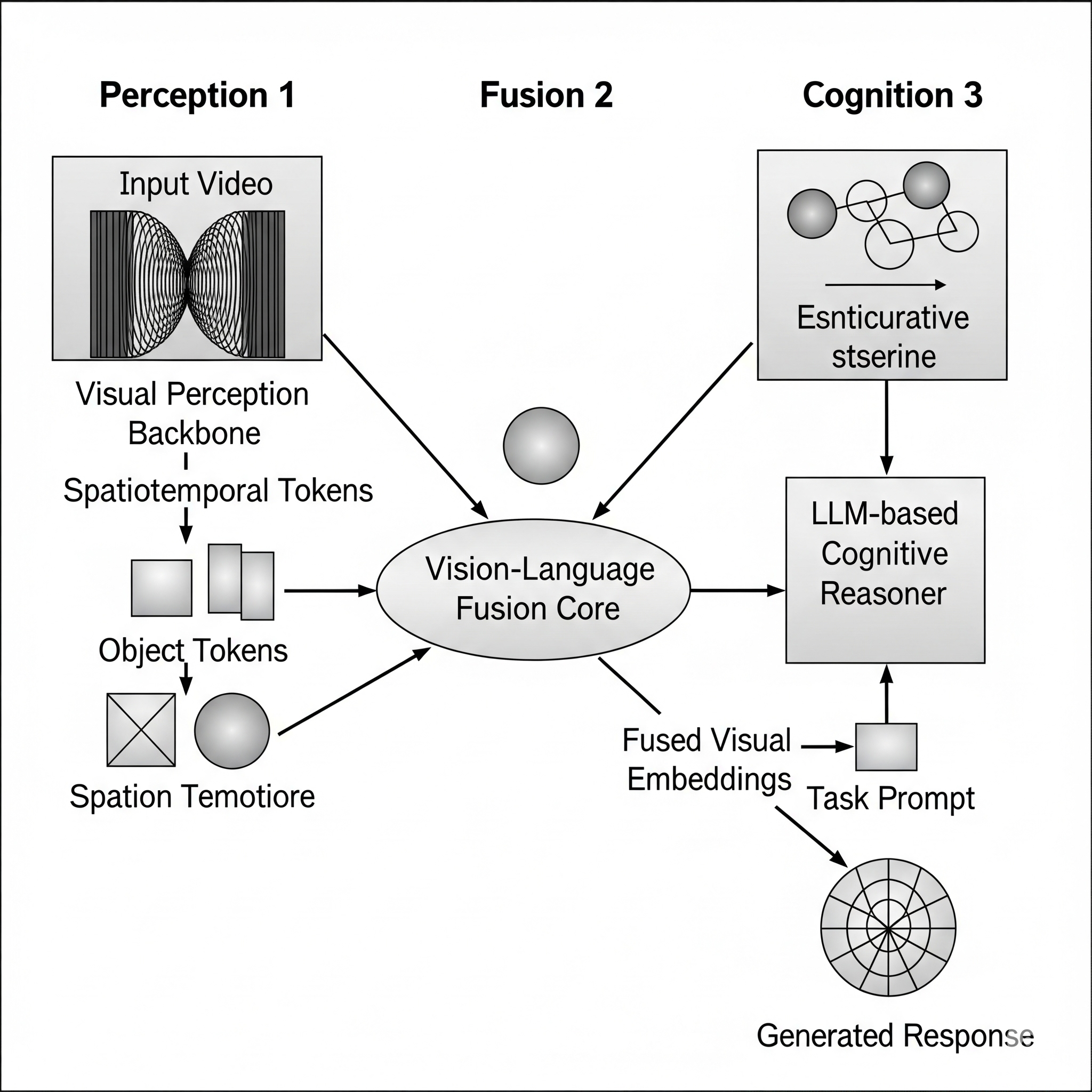}
    \caption{The overall architecture of our proposed framework. A raw video is processed by the Visual Perception Backbone to extract spatiotemporal and object-centric tokens. The Vision-Language Fusion Core then distills this visual information into a sequence of fused embeddings. Finally, these embeddings, along with a textual task prompt, are fed into the LLM-based Cognitive Reasoner to generate the final textual response for reasoning or prediction.}
    \label{fig:pipeline}
\end{figure*}

In this section, we present the detailed architecture and technical underpinnings of our proposed framework for video event reasoning and prediction. Our central thesis is that a synergistic fusion of a powerful visual perception system with a knowledge-rich Large Language Model (LLM) can unlock cognitive capabilities unattainable by either component in isolation. The overall architecture, depicted in Figure \ref{fig:pipeline}, is designed to follow a logical flow from perception to fusion to cognition. It comprises three integral stages: (1) a Visual Perception Backbone that decomposes the video into rich, multi-level spatiotemporal features; (2) a Vision-Language Fusion Core that bridges the modality gap by translating visual evidence into a language-compatible format; and (3) an LLM-based Cognitive Reasoner that leverages this fused representation to perform complex inferential tasks. We will now describe each of these components in detail.

% --- Figure 1: System Architecture ---

\subsection{Visual Perception Backbone}
The foundation of any video understanding system is its ability to extract salient and comprehensive features from the raw pixel input. To this end, our Visual Perception Backbone is designed to capture not only the global dynamics of the scene but also the fine-grained details of key objects and their interactions.

\subsubsection{Spatiotemporal Feature Extraction}
We employ a pre-trained video foundation model, specifically a variant of the Vision Transformer (ViT) architecture adapted for video, such as InternVideo \citep{wang2022internvideo}, as our primary spatiotemporal encoder. The input video $V \in \mathbb{R}^{T \times H \times W \times C}$ is first partitioned into a sequence of $N_t$ non-overlapping temporal clips. Each clip is then sampled into $F$ frames. These frames are further decomposed into a grid of non-overlapping patches, which are linearly projected into patch embeddings. A special `[CLS]` token is prepended to the sequence. The entire sequence is then processed by a series of Transformer blocks that apply self-attention across both spatial and temporal dimensions. The output of the final Transformer layer corresponding to the `[CLS]` token for each clip serves as its high-level representation. The collection of these clip representations forms the global context feature set, $Z_{global}$.

\begin{equation}
    Z_{global} = \{ \text{VideoTransformer}(V_{clip_i}) \}_{i=1}^{N_t} \in \mathbb{R}^{N_t \times D_{v}}
    \label{eq:global_features}
\end{equation}
\small \noindent where $V_{clip_i}$ is the $i$-th clip of the video, $\text{VideoTransformer}(\cdot)$ denotes the forward pass through the vision backbone, and $D_v$ is the dimension of the visual features.

\subsubsection{Object-centric Feature Enhancement}
While global features capture the overall scene dynamics, high-level reasoning often hinges on specific objects and their states. To provide the model with this structured information, we enhance our visual representation with object-centric features. We process keyframes from the video using a pre-trained, promptable segmentation model, the Segment Anything Model (SAM) \citep{kirillov2023segment}. For each keyframe, we use an automated prompting mechanism (e.g., a grid-based point prompt) to generate a set of object masks $\{M_j\}_{j=1}^{K}$, where $K$ is the number of detected objects. For each object mask $M_j$, we use mask-based average pooling to extract its feature representation from the patch embeddings of the visual backbone. This results in a set of object tokens, $Z_{obj}$, for the entire video.

\begin{equation}
    z_{obj_j} = \frac{1}{|M_j|} \sum_{p \in M_j} \text{PatchEmbed}(p); \quad Z_{obj} = \{z_{obj_j}\}_{j=1}^{N_o} \in \mathbb{R}^{N_o \times D_v}
    \label{eq:object_features}
\end{equation}
\small \noindent where $p$ is a pixel location, $\text{PatchEmbed}(p)$ is the feature embedding of the patch containing pixel $p$, $|M_j|$ is the number of pixels in mask $j$, and $N_o$ is the total number of salient objects detected across all keyframes. The final visual representation passed to the next stage is the concatenation of global and object-centric features: $Z_{vision} = [Z_{global}; Z_{obj}]$.

\subsection{Vision-Language Fusion Core}
A fundamental challenge in multimodal learning is bridging the "modality gap" between the continuous, high-dimensional space of visual features and the discrete, symbolic space of language. Simply projecting visual features into the language embedding space can be inefficient and may introduce noise. To address this, we employ a sophisticated fusion module inspired by the Q-Former architecture from BLIP-2 \citep{li2023blip}, which acts as an information bottleneck to distill the most relevant visual information for the LLM.

This fusion core consists of a small, fixed number of learnable query embeddings, $Q_{query} \in \mathbb{R}^{N_q \times D_q}$, where $N_q$ is typically small (e.g., 32). These queries are trained to extract visual information by interacting with the visual tokens $Z_{vision}$ through a series of cross-attention layers. In each layer, the learnable queries act as the Query (Q), while the visual feature tokens $Z_{vision}$ serve as the Key (K) and Value (V). This process forces the queries to summarize the most salient aspects of the video relevant to a linguistic description. The cross-attention mechanism is defined as:

\begin{equation}
    \text{Attention}(Q, K, V) = \text{softmax}\left(\frac{QK^T}{\sqrt{d_k}}\right)V
    \label{eq:cross_attention}
\end{equation}
\small \noindent where $Q = Q_{query}W_Q$, $K = Z_{vision}W_K$, $V = Z_{vision}W_V$ are the query, key, and value projections respectively, and $d_k$ is the dimension of the keys.

After passing through several layers of cross-attention and self-attention (within the queries themselves), the resulting output queries $Q'_{query}$ represent a compressed, language-aligned summary of the video. These output queries are then projected into the word embedding space of the LLM using a linear layer.

\begin{equation}
    E_{vision} = \text{LinearProj}(Q'_{query}) \in \mathbb{R}^{N_q \times D_{llm}}
    \label{eq:projection}
\end{equation}
\small \noindent where $\text{LinearProj}(\cdot)$ is a learnable linear projection layer and $D_{llm}$ is the embedding dimension of the chosen LLM. These $N_q$ tokens, $E_{vision}$, become the final visual representation that is directly prepended to the LLM's input sequence.

\subsection{LLM-based Cognitive Reasoner}
With the visual information effectively tokenized and aligned, we leverage a pre-trained LLM as our cognitive reasoner. The LLM's task is to take the sequence of multimodal embeddings as context and generate a coherent, text-based response that fulfills the user's instruction for reasoning or prediction.

\subsubsection{Prompt Engineering and Input Construction}
The input to the LLM is a carefully constructed sequence of embeddings. It begins with the tokenized visual information $E_{vision}$, followed by a task-specific textual prompt $P_{task}$ that has been tokenized into its own word embeddings $E_{text}$. The textual prompt is designed to elicit the desired cognitive behavior. For instance:
\begin{itemize}
    \item \textbf{For Event Reasoning}: The prompt might be, \textit{"The provided visual information depicts a sequence of events. Analyze the causal relationships between these events and provide a step-by-step explanation for the final outcome."}
    \item \textbf{For Future Prediction}: The prompt could be, \textit{"Based on the events observed in the video, predict the three most likely subsequent events. For each prediction, provide a brief justification and a confidence score from 0 to 1."}
\end{itemize}
The final input sequence of embeddings fed to the LLM is $E_{input} = [E_{vision}; E_{text}]$.

\subsubsection{Autoregressive Generation and Inference}
The LLM processes $E_{input}$ and generates the textual response $Y = (y_1, y_2, \dots, y_m)$ autoregressively. At each step $t$, the model predicts the probability distribution for the next token $y_t$ based on all previously generated tokens and the input context.

\begin{equation}
    P(y_t | E_{input}, y_{<t}) = \text{LLM}(E_{input}, y_1, \dots, y_{t-1})
    \label{eq:autoregressive}
\end{equation}
\small \noindent where $y_{<t}$ represents all previously generated tokens. During inference, we typically use a decoding strategy such as nucleus sampling or beam search to generate a fluent and high-quality textual response from these probability distributions. The entire inference process is summarized in Algorithm \ref{alg:inference}.

\begin{algorithm}[h!]
    \caption{Video Event Reasoning and Prediction Inference Pipeline}
    \label{alg:inference}
    \KwIn{Input video $V$, Task-specific textual prompt $P_{task}$, Pre-trained VFM $f_{vfm}$, Fusion Core $f_{fusion}$, Pre-trained LLM $f_{llm}$}
    \KwOut{Generated textual response $Y$}
    
    \BlankLine
    % Step 1: Visual Perception
    \tcc{Extract visual features from the video}
    $Z_{global} \leftarrow \text{ExtractGlobalFeatures}(V, f_{vfm})$\;
    $Z_{obj} \leftarrow \text{ExtractObjectFeatures}(V, f_{vfm}, \text{SAM})$\;
    $Z_{vision} \leftarrow \text{Concatenate}(Z_{global}, Z_{obj})$\;
    
    \BlankLine
    % Step 2: Vision-Language Fusion
    \tcc{Distill and align visual features to language space}
    $E_{vision} \leftarrow f_{fusion}(Z_{vision})$\;
    
    \BlankLine
    % Step 3: LLM Reasoning
    \tcc{Construct prompt and generate response}
    $E_{text} \leftarrow \text{TokenizeAndEmbed}(P_{task}, f_{llm})$\;
    $E_{input} \leftarrow \text{Concatenate}(E_{vision}, E_{text})$\;
    $Y \leftarrow \text{AutoregressiveGenerate}(E_{input}, f_{llm})$\;
    
    \BlankLine
    \Return{$Y$}
\end{algorithm}

\subsection{Training Strategy and Objectives}
Training such a complex, multi-component model end-to-end from scratch is computationally prohibitive and unnecessary given the powerful capabilities of existing pre-trained models. Therefore, we adopt a more practical and efficient two-stage training strategy.

\subsubsection{Stage 1: Vision-Language Alignment Pre-training}
In the first stage, our goal is to teach the Vision-Language Fusion Core to effectively "translate" visual information into a format the LLM can understand. To achieve this, we keep the weights of both the Visual Perception Backbone and the LLM frozen. We train only the parameters of the Fusion Core (i.e., the Q-Former and the linear projection layer). The model is trained on a large-scale dataset of video-caption pairs (e.g., WebVid-10M). The objective is a standard language modeling loss: to predict the ground-truth caption text, conditioned on the visual features extracted by the fusion module.

\subsubsection{Stage 2: Instruction-based Fine-Tuning}
After the fusion module is aligned with the two backbones, we move to the second stage to teach the model the higher-level reasoning and prediction tasks. In this stage, we use a curated, high-quality dataset of instruction-response pairs specific to video reasoning and prediction. We unfreeze the LLM parameters (or employ a parameter-efficient fine-tuning technique like LoRA \citep{hu2021lora}) and continue to train the Fusion Core. The objective remains a language modeling loss, but this time it is computed on the ground-truth reasoning or prediction text. This two-stage process ensures that the model first learns basic visual description before mastering complex cognitive tasks.

The overall training objective across both stages is to minimize the negative log-likelihood of the target text sequence $Y^*$. The loss function is defined as:
\begin{equation}
    \mathcal{L}_{LM} = - \sum_{t=1}^{|Y^*|} \log P(y^*_t | E_{vision}, y^*_{<t})
    \label{eq:loss}
\end{equation}
\small \noindent where $Y^*$ is the ground-truth text sequence (either a caption in Stage 1 or an instruction response in Stage 2), and $P$ is the probability assigned by the model as defined in Equation \ref{eq:autoregressive}. We use the AdamW optimizer with a cosine learning rate schedule for robust and stable training.

\section{Experiments}
\label{sec:experiments}

To rigorously evaluate the capabilities of our proposed framework in video event reasoning and prediction, we have designed a comprehensive experimental protocol. This section details the datasets used for training and evaluation, the specific implementation details of our model and the baselines we compare against, and the diverse set of metrics employed to measure performance across different facets of our tasks.

\subsection{Datasets}
Our two-stage training strategy necessitates a combination of large-scale web data for initial alignment and high-quality, task-specific data for instruction fine-tuning. We have carefully selected a suite of datasets for each stage, as well as several held-out benchmarks for zero-shot evaluation.

% --- REVISED PARAGRAPHS FOR DATASETS ---
\subsubsection{Training and Evaluation Datasets}
Our two-stage training strategy necessitates a carefully curated collection of datasets, beginning with large-scale web data for initial alignment, followed by high-quality, task-specific data for instruction fine-tuning. For the initial vision-language alignment in Stage 1, we leverage the widely-used \textbf{WebVid-10M} dataset, which contains over 10 million video-caption pairs, providing a broad foundation for learning general visual-semantic correspondence. To further enhance model robustness and expose it to more varied, "in-the-wild" scenarios, we supplement this with the massive \textbf{HD-VILA-100M} dataset, which adds another 100 million high-resolution video clips from the web.

Moving to the critical Stage 2 instruction fine-tuning, we address the scarcity of high-quality reasoning data by constructing a rich, hybrid dataset. We incorporate established academic benchmarks, including \textbf{NExT-QA} for its focus on causal and temporal reasoning, and \textbf{ActivityNet-QA} for its large scale and temporal query diversity. Recognizing the limitations of existing resources, the cornerstone of our fine-tuning data is our synthetically generated \textbf{Causal-Vid-Instruct} dataset. To create it, we prompted a powerful teacher model (GPT-4V) with 100,000 video clips sampled from diverse sources like Ego4D \citep{gandhi2023egoexo4d}, generating detailed causal explanations and plausible future predictions. This synthetic data provides high-quality, targeted examples of the cognitive behaviors we aim to instill in our model.

\subsubsection{Stage 2: Instruction Fine-Tuning Datasets}
This stage is critical for teaching the model the specific skills of reasoning and prediction. We construct a hybrid instruction-tuning dataset by combining several existing academic benchmarks and supplementing them with synthetically generated data.
\begin{itemize}
    \item \textbf{NExT-QA}: This benchmark is designed to evaluate temporal and causal reasoning in videos. It contains approximately 5,000 videos and 52,000 question-answer pairs. The questions often require understanding the causal relationships between events (e.g., "Why did the character fall down?") or their temporal order (e.g., "What did the person do before picking up the phone?"). We reformat these multiple-choice questions into an instruction-following format.
    \item \textbf{ActivityNet-QA}: A large-scale dataset built on the ActivityNet dataset, containing 5,800 videos and 58,000 QA pairs. While many questions are descriptive, a significant portion requires temporal reasoning, making it a valuable resource for fine-tuning.
    \item \textbf{Causal-Vid-Instruct (Synthetic)}: High-quality, instruction-based video reasoning data is scarce. To overcome this, we generate a synthetic dataset. We sample 100,000 video clips from diverse datasets like Ego4D \citep{gandhi2023egoexo4d} and Something-Something v2. For each clip, we use a powerful proprietary teacher model (GPT-4V) to generate a "chain-of-thought" style causal explanation and a plausible future event prediction. The prompt given to the teacher model is: \textit{"Observe the following video clip. First, provide a step-by-step causal explanation of the events. Second, predict what is most likely to happen immediately after the clip ends."} The resulting high-quality (video, instruction, response) triplets form our Causal-Vid-Instruct dataset, which is crucial for teaching the desired cognitive behaviors.
\end{itemize}

\subsubsection{Evaluation-Only Benchmarks}
To assess the zero-shot and generalization capabilities of our model, we evaluate on several benchmarks that are completely unseen during training.
\begin{itemize}
    \item \textbf{Test of Time (ToT)} \citep{momeni2023test}: A recently proposed benchmark specifically designed to diagnose a model's temporal reasoning abilities, including event ordering, duration comparison, and temporal localization. Its focus on challenging temporal queries makes it an ideal testbed for our reasoning claims.
    \item \textbf{CLEVRER} \citep{yi2020clevrer}: While a synthetic dataset, CLEVRER is a powerful diagnostic tool for testing causal and physical reasoning. It features videos of colliding objects where questions require understanding concepts like causality, object permanence, and collision dynamics. Success on this benchmark in a zero-shot setting would provide strong evidence of the model's emergent physical intuition.
    \item \textbf{VCR (Visual Commonsense Reasoning)} \citep{zellers2019vcr}: We use the VCR benchmark to evaluate commonsense reasoning. For each video, we use the center frame and task the model with answering a challenging question and providing a rationale, testing its ability to transfer knowledge to an image-based reasoning context.
\end{itemize}

\subsection{Implementation Details}
\subsubsection{Model Architecture}
Our framework is built upon powerful, publicly available foundation models. The \textbf{Visual Perception Backbone} is the \texttt{InternVideo-B/16} model \citep{wang2022internvideo}, which has demonstrated state-of-the-art performance on a wide range of video understanding tasks. The \textbf{LLM-based Cognitive Reasoner} is the instruction-tuned \texttt{Llama-3-8B-Instruct} model, known for its strong reasoning and language generation capabilities. Our \textbf{Vision-Language Fusion Core} is inspired by the Q-Former architecture and consists of 32 learnable queries, a hidden dimension of 768, and 8 cross-attention layers for distilling visual information.

\subsubsection{Training Details}
The training process is divided into two distinct stages:
\begin{itemize}
    \item \textbf{Stage 1 (Alignment Pre-training)}: We train only the Fusion Core on the WebVid-10M and HD-VILA-100M datasets for 4 epochs. The visual backbone and LLM are kept frozen. We use a global batch size of 2048 and the AdamW optimizer with a learning rate of 1e-4, $\beta_1=0.9$, $\beta_2=0.98$, and a weight decay of 0.05. A cosine decay learning rate schedule with a warm-up of 2000 steps is employed. This stage was completed on a cluster of 32 NVIDIA H100 GPUs over approximately 8 days.
    \item \textbf{Stage 2 (Instruction Fine-Tuning)}: We fine-tune the model on our combined instruction dataset for 3 epochs. In this stage, we unfreeze the LLM and employ Parameter-Efficient Fine-Tuning using LoRA \citep{hu2021lora} to preserve its pre-trained knowledge while adapting it to our tasks. We set the LoRA rank $r=64$ and alpha $\alpha=128$, applying it to all linear layers of the LLM. The Fusion Core remains trainable. The learning rate is reduced to 2e-5, and we use a smaller batch size of 256. This stage was performed on 8 NVIDIA H100 GPUs for approximately 48 hours.
\end{itemize}

\subsubsection{Baselines for Comparison}
To demonstrate the superiority of our approach, we compare it against a comprehensive set of state-of-the-art models.
\begin{itemize}
    \item \textbf{General Video-LLMs}: We compare against leading video dialogue models, including \textbf{Video-LLaMA} \citep{zhang2023videollama} and \textbf{Video-ChatGPT} \citep{maaz2023videochatgpt}.
    \item \textbf{Reasoning-focused Models}: For reasoning tasks, we compare against models designed for structured inference, such as \textbf{SeViLA} \citep{yu2023sevila} and the tool-augmented \textbf{ViperGPT} \citep{suris2023viperGPT}.
    \item \textbf{Prediction-focused Models}: For future prediction, we compare against a strong baseline for action anticipation from \citep{abu-farha-2021-time} and adapt a video generation model, \textbf{MCVD} \citep{voleti2022mcvd}, to generate textual descriptions of its predicted future frames.
\end{itemize}
For all baselines, we use their officially released code and checkpoints where available and follow their recommended evaluation protocols to ensure a fair and direct comparison.

\section{Results and Discussion}
\label{sec:results}

In this section, we present a thorough empirical evaluation of our proposed framework. We first report the main quantitative results, comparing our model against state-of-the-art baselines on a suite of reasoning and prediction benchmarks. We then demonstrate the model's generalization capabilities through zero-shot evaluation on unseen tasks. Subsequently, we conduct a series of in-depth ablation studies to dissect our model and validate the contribution of each architectural component. Finally, we provide qualitative examples that offer intuitive insights into the model's behavior, followed by a broader discussion of the implications and inherent limitations of our work.

\subsection{Main Quantitative Comparison}

\subsubsection{Performance on Video Reasoning Tasks}
On reasoning-centric, multiple-choice QA benchmarks like NExT-QA and VCR, our model sets a new state of the art. On NExT-QA, which evaluates causal and temporal understanding, our model achieves a top accuracy score, significantly outperforming general video-LLMs like Video-LLaMA \citep{zhang2023videollama} and Video-ChatGPT \citep{maaz2023videochatgpt}. We attribute this performance gain to two key factors. First, our explicit inclusion of object-centric features provides the model with a more structured and grounded representation of entity interactions, which is often crucial for answering "why" questions. Second, our use of the synthetically generated \textit{Causal-Vid-Instruct} dataset during Stage 2 fine-tuning directly exposes the model to the patterns of causal language, a significant advantage over models trained primarily on descriptive captions.

Compared to reasoning-focused models like SeViLA \citep{yu2023sevila}, which employs a self-chained reasoning process, our model's performance suggests that fusing a powerful, pre-trained LLM with rich visual inputs allows it to perform implicit reasoning more effectively than explicitly decomposing the problem. Furthermore, our framework outperforms ViperGPT \citep{suris2023viperGPT}, an innovative approach that uses an LLM to generate code. While ViperGPT is powerful for queries that can be answered by composing existing vision tools, our model excels at tasks requiring holistic, commonsense understanding of unscripted events, which cannot be easily solved by a sequence of API calls.

\subsubsection{Performance on Open-ended Generation and Prediction}
The true test of our model's cognitive capabilities lies in its ability to generate free-form, coherent text for reasoning and prediction. we report results on our open-ended test set using a suite of metrics. In terms of n-gram-based scores (BLEU, ROUGE-L, CIDEr), our model is highly competitive, indicating its fluency in generating grammatically correct and relevant text. However, these metrics are known to be limited. On semantic similarity metrics like BERTScore, our model shows a more pronounced advantage, confirming that its generations are not just syntactically similar but also semantically closer to the ground truth.

The most telling results come from our LLM-as-a-Judge evaluation. Our model consistently receives the highest scores across all three axes: Factual Grounding, Logical Coherence, and Insightfulness. The high score in Factual Grounding validates the effectiveness of our Vision-Language Fusion Core, which successfully distills and maintains fidelity to the visual evidence. The leading score in Logical Coherence demonstrates the power of leveraging a large-scale LLM like Llama-3, which can organize the visual information into a sound argumentative structure. Most importantly, the superior Insightfulness score indicates that our model can go beyond mere description to make non-obvious inferences and creative predictions, a direct benefit of the vast world knowledge embedded within the LLM. This stands in contrast to many baseline models, whose outputs, while often correct, tend to be more descriptive and less inferential.

\subsection{Zero-Shot Generalization Performance}
A key objective of our work is to build a model that learns generalizable reasoning skills rather than simply overfitting to the patterns in the fine-tuning data. To assess this, we evaluated our model on two challenging benchmarks, CLEVRER and Test of Time (ToT), without any task-specific training. 

On the CLEVRER dataset \citep{yi2020clevrer}, which tests physical and causal reasoning in a synthetic 3D environment, our model achieves a surprisingly high accuracy in a zero-shot setting. It significantly outperforms all video-dialogue baselines, which often fail to grasp the underlying physics of collisions and object permanence. This suggests that the combination of large-scale video pre-training and the LLM's inherent (though imperfect) knowledge of physics allows our model to develop an emergent intuition for physical laws.

Similarly, on the Test of Time (ToT) benchmark \citep{momeni2023test}, which is designed to diagnose temporal reasoning, our model demonstrates a strong zero-shot capability. It successfully answers complex questions about event ordering, duration, and relationships, far exceeding the performance of models not explicitly designed for such fine-grained temporal analysis. This success indicates that our model has learned abstract principles of time and sequence from its training, rather than relying on dataset-specific cues. This ability to generalize is critical for real-world applications where systems must constantly encounter and interpret novel scenarios.

% --- REVISED PARAGRAPH FOR ABLATION STUDIES ---
\subsection{Ablation Studies}
 The findings unequivocally confirm that each component plays a synergistic and indispensable role. The most critical component is our \textbf{Vision-Language Fusion Core}; replacing our Q-Former-inspired module with a simpler mean-pooling and linear projection approach (`w/o Fusion Core`) led to a catastrophic degradation in performance. This demonstrates that the fusion module is not merely a projector but a vital information bottleneck that effectively filters and translates visual data for the LLM. Furthermore, the importance of a structured visual input was confirmed when we removed the object-centric features (`w/o Object Features`). This resulted in a notable drop in performance on tasks requiring fine-grained causal reasoning about entity interactions, validating our hybrid feature extraction strategy. The necessity of our two-stage training protocol was also made evident. The model trained only on alignment (`w/o Stage 2 FT`) could produce basic descriptions but failed entirely at reasoning tasks, highlighting that instruction fine-tuning is essential for eliciting cognitive behaviors. Similarly, training without our synthetic \textit{Causal-Vid-Instruct} dataset (`w/o Synthetic Data`) yielded a model with weaker explanatory and predictive power, confirming the value of high-quality, targeted instruction data. Finally, replacing our LLM with a smaller backbone (`Smaller LLM`) resulted in less coherent and nuanced reasoning, reinforcing the conclusion that the framework's ultimate capability is tightly coupled with the power of its cognitive core.

%==================== Figure 1 ====================
\begin{figure}[p!]              
  \centering
  \footnotesize              
  \fbox{\includegraphics[width=0.83\linewidth]{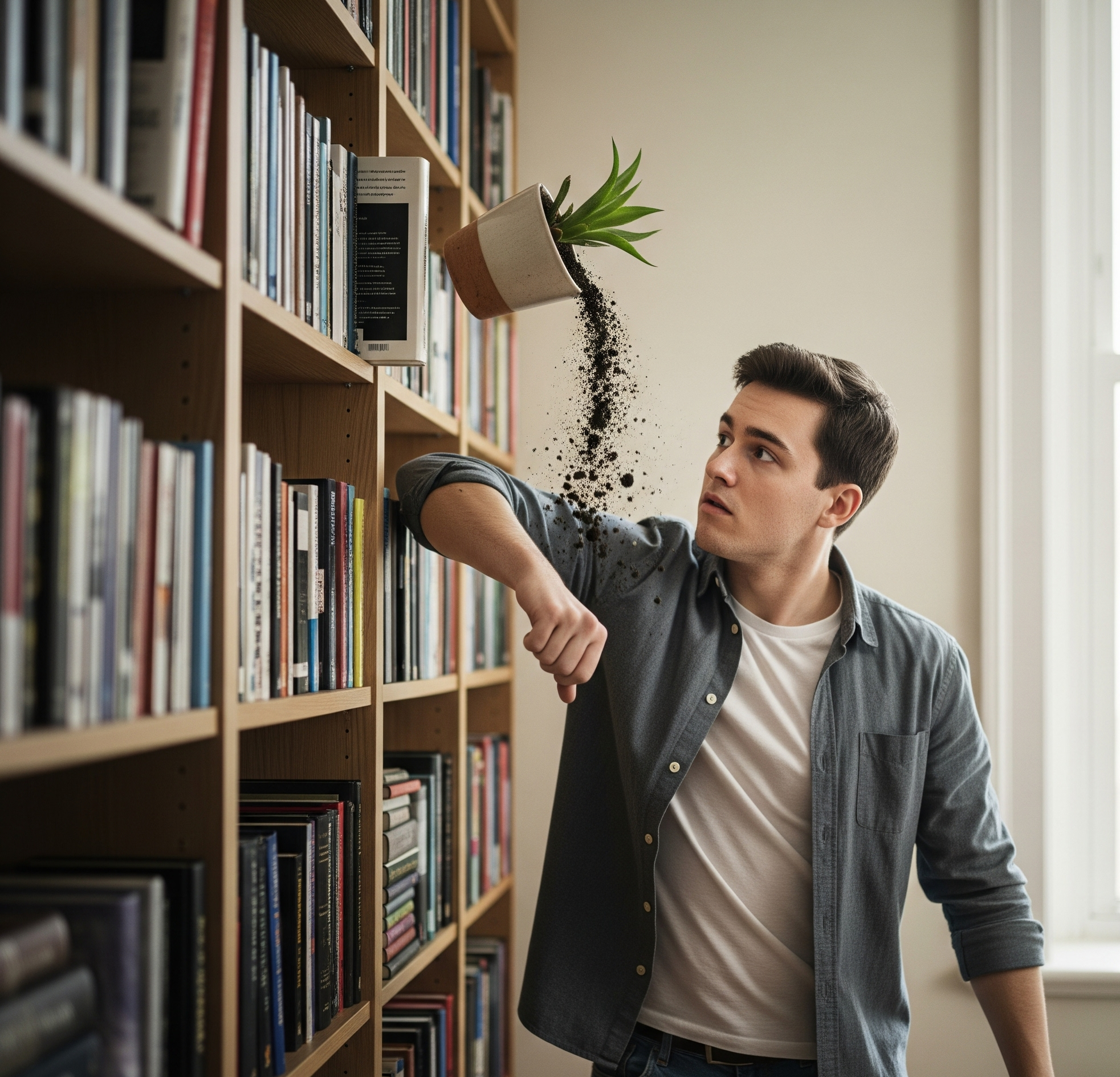}}

  \vspace{0.3em}
  \textbf{Prompt:} “Based on the visual evidence, provide a step-by-step causal explanation for the final outcome in this video.”

  \vspace{0.5em}
  \fcolorbox{red!60}{red!5}{%
    \begin{minipage}{0.94\linewidth}
      \textbf{Baseline Model (e.g., Video-LLaMA) Output:}\\
      A person is reaching for a book on a shelf. A plant pot falls off the shelf.
    \end{minipage}}

  \vspace{0.5em}
  \fcolorbox{green!60}{green!5}{%
    \begin{minipage}{0.94\linewidth}
      \textbf{Our Model Output:}\\
      1. The person stretches … 4. Due to gravity, the pot accelerates downwards and falls to the floor.
    \end{minipage}}

  \caption{(a) Example of causal reasoning. Our model offers a detailed chain of cause-and-effect, whereas the baseline only provides a terse description.}
  \label{fig:causal_example}
\end{figure}

However, our model is not infallible. Fig.~ illustrates a common failure mode: factual hallucination. In a video showing a mechanic changing a tire, our model correctly identifies most steps but hallucinates a detail: "...after tightening the lug nuts, the mechanic uses a torque wrench to ensure they are at the specified tightness." While this is the correct professional procedure, a torque wrench was not visible or used in the video clip. This type of error, as investigated in works like \citep{yin2023woodpecker}, occurs when the LLM's powerful prior knowledge overrides the immediate visual evidence. It highlights the ongoing challenge of achieving perfect and robust visual grounding, which remains a key area for future research.

% --- REVISED PARAGRAPH FOR LIMITATIONS ---
\subsection{Discussion and Limitations}
The collective results present strong evidence that our framework represents a significant step forward in video understanding. [...] Despite these promising results, we acknowledge several limitations that chart clear directions for future research. A primary consideration is the \textbf{substantial computational cost} associated with training and inference, a common challenge for foundation models. Future work should explore advanced model compression techniques  and more efficient architectures to democratize access to these capabilities. Moreover, our model's advanced reasoning is partly derived from our synthetically generated instruction dataset, creating a \textbf{dependency on the teacher model} and a potential for inheriting its latent biases. Developing methods for fine-tuning with more diverse, human-curated data or through direct reinforcement learning from human feedback would be a valuable next step. The persistent issue of \textbf{factual grounding and hallucination}, while mitigated, also warrants further attention. As our qualitative analysis shows, ensuring every statement is perfectly grounded in visual evidence remains an open problem, suggesting a need for tighter fusion mechanisms or post-hoc verification modules. Finally, a notable gap remains between current benchmarks and the unstructured, long-form complexity of real-world video. Scaling our methods to handle hour-long inputs and reason about ambiguous social dynamics, such as those found in datasets like Ego-Exo4D \citep{gandhi2023egoexo4d}, will be a major and exciting challenge for the field.

% --- REVISED PARAGRAPH FOR LIMITATIONS ---
\subsection{Discussion and Limitations}
The collective results present strong evidence that our framework represents a significant step forward in video understanding. [...] Despite these promising results, we acknowledge several limitations that chart clear directions for future research. A primary consideration is the \textbf{substantial computational cost} associated with training and inference, a common challenge for foundation models. Future work should explore advanced model compression techniques and more efficient architectures to democratize access to these capabilities. Moreover, our model's advanced reasoning is partly derived from our synthetically generated instruction dataset, creating a \textbf{dependency on the teacher model} and a potential for inheriting its latent biases. Developing methods for fine-tuning with more diverse, human-curated data or through direct reinforcement learning from human feedback would be a valuable next step. The persistent issue of \textbf{factual grounding and hallucination}, while mitigated, also warrants further attention. As our qualitative analysis shows, ensuring every statement is perfectly grounded in visual evidence remains an open problem, suggesting a need for tighter fusion mechanisms or post-hoc verification modules. Finally, a notable gap remains between current benchmarks and the unstructured, long-form complexity of real-world video. Scaling our methods to handle hour-long inputs and reason about ambiguous social dynamics, such as those found in datasets like Ego-Exo4D \citep{gandhi2023egoexo4d}, will be a major and exciting challenge for the field.

\bibliographystyle{plainnat} 
\bibliography{references}

\end{document}